\documentclass[11pt,a4paper]{article}
\usepackage[hyperref]{stylefiles/emnlp-ijcnlp-2019}
\usepackage{times}
\usepackage{latexsym}
\usepackage{subfig}
\usepackage{graphicx}
\usepackage{booktabs}
\usepackage{url}

\aclfinalcopy % Uncomment this line for the final submission

\newcommand{\todo}[1]{}

\title{The Fluidity of Concept Representations in Human Brain Signals}

\author{Eva Hendrikx \\
  University of Amsterdam \\
  {\tt evi.hendrikx@student.uva.nl} \\\And
  Lisa Beinborn \\
  University of Amsterdam \\
  {\tt l.beinborn@uva.nl} \\}

\date{}

\begin{document}
\maketitle
\begin{abstract}
Cognitive theories of human language processing often distinguish between concrete and abstract concepts. In this work, we analyze the discriminability of concrete and abstract concepts in fMRI data using a range of analysis methods. We find that the distinction can be decoded from the signal with an accuracy significantly above chance, but it is not found to be a relevant structuring factor in clustering and relational analyses. From our detailed comparison, we obtain the impression that human concept representations are more fluid than dichotomous categories can capture. We argue that fluid concept representations lead to more realistic models of human language processing because they better capture the ambiguity and underspecification present in natural language use. 

\end{abstract}
%\todo{If space and time, describe methods more formally.}

% A PAPER IS JUST A PROGRESS REPORT ;-).
\section{Introduction}
Language researchers often group words into categories. Lexicographers categorize words by their syntactic category, historical linguists categorize them by their ancestors, computational linguists categorize by frequency, and psycholinguists distinguish words by categories such as concreteness, imageability, meaningfulness, and age of acquisition \cite{MRCdatabase}. 

The distributional word representations that are most commonly used by computer scientists nowadays are high-dimensional vectors. The dimensions of these vectors cannot easily be interpreted as linguistic categories \cite{lewis2013}. Relations between words are instead characterized by their representational similarity, which is measured as proximity in the vector space. Several gold standard datasets for similarity assessment exist and provide a fundamental resource for the development and evaluation of computational word representations. Limitations of similarity judgments are low inter-dataset and inter-annotator agreement \cite{batchkarov2016}. This arises because similarity between words is a highly fluid concept that ranges over multiple properties (e.g., shape, semantic category, emotion associated with the word). 

% I put this back in because I like the contrast. People try hard to map distributional representations back into dichotomous categories while we argue that psycholinguists should accept that the relations between concepts are more finegrained and fluid. It also provides a link to the similarity paragraph before.  
In cognitive science research, psycholinguistic categories play an important role and the low interpretability of distributional representations can pose a challenge for interdisciplinary research. Therefore, recent work by \newcite{faruqui2015sparse} aims to project distributional representations back onto interpretable linguistic dimensions.  
% Not sure if the verb/noun distinction is the best example here, but we can leave it for the moment
To which extent these dichotomous linguistic categories of words are reflected in the human brain remains a topic of debate, as various studies come to different conclusions. For example, \newcite{rapp2002} claim that patient data convincingly shows different underlying representations for verbs and nouns, whereas \newcite{bird2003} find no differences in processing, when controlling for confounding factors, such as imageability.

%In this work, we focus on the concreteness category because it has been vividly discussed in light of the embodiment debate and it has been assumed that concrete and abstract concepts are represented differently
% In this work, we focus on the question whether concrete and abstract concepts are represented differently in the brain. The dichotomy of abstract and concrete concepts has been vividly discussed in light of the embodiment debate  and different underlying representations in the brain are often assumed.

% I changed this slightly again, okay for you? Yes, great!
In this work, we focus on the concreteness category because it has been vividly discussed in light of the embodiment debate \cite{barsalou2008,pecher2011}. It is often assumed that concrete and abstract words are represented differently in the brain, but topological analyses have not yet reached consensus on the involved brain areas  \cite{mestres2009,tettamanti2008,wallentin2005,wise2000}. We use brain activation data from fMRI analyses because they provide high spatial resolution. Previous studies which examined linking hypotheses between computational representations and fMRI patterns of concrete and abstract words yielded contradictory results \cite{anderson2017, bulat2017}. We aim at consolidating their analyses by using multiple context paradigms. 

Previous work indicates that the interpretation of fMRI data strongly depends on the analysis method and the evaluation metrics even within a consistent experimental paradigm \cite{beinborn2019}. To increase the transparency of our results we apply a range of computational analysis methods %to inspect whether concrete and abstract words are represented differently in the brain 
and make our code publicly available at \url{https://github.com/evi-hendrikx/Fluid_Concept_Rep}. Our analysis of the concreteness category is the first to examine relational effects of various computational representations for different context paradigms (to the best of our knowledge). Furthermore, we compare data-driven searchlight analysis with more traditional region of interest selections.

\section{Computational Analysis of Cognitive Concept Representations}
We briefly sketch previous computational analyses with cognitive data and then focus on the distinction between concrete and abstract concepts. 
\subsection{Analyzing Cognitive Data}
Computational analyses of cognitive data provide an interdisciplinary bridge between computational and cognitive models of language. On the one hand, experimental data such as response times or subjective ratings have a strong influence on the development and evaluation of computational models of language \cite{monsalve2012,Resnik2010}. Eye-tracking measures are used to investigate human attention and guide the development of models for sentence understanding \cite{Barrett2018}, sentiment analysis, \cite{hollenstein2019advancing} and multi-modal processing \cite{sugano2016seeing}. Fine-grained syntactic and semantic processing is often modeled using electroencephalography data \cite{Hale2018,Fyshe2016,Frank2013,Sudre2012}. 

On the other hand, cognitive researchers conduct computational analyses to detect patterns in experimental data. This is particularly important when dealing with high-dimensional data from magnetoencephalography and functional magnetic resonance imaging (fMRI) scans to study the functional localization of language processing. We focus on fMRI data in this paper and distinguish between two types of computational analyses: discriminative and relational analyses.

Discriminative analyses investigate whether it is possible to group the representational patterns into classes. A popular discriminative analysis is known as decoding. For this task, a computational model learns to discriminate the fMRI patterns for different linguistic categories, e.g., abstract and concrete concepts \cite{anderson2017},
various syntactic classes \cite{Bingel2016,Li2018}, or levels of syntactic complexity \cite{Brennan2016}. Another discriminative analysis is clustering. 
For clustering analyses, the categories of interest are not defined a priori, but the data is automatically grouped according to shared characteristics in the representations.

Relational analyses aim at establishing a link between the fMRI signal and a computational representation of the stimulus. 
The results by \newcite{Mitchell2008} indicate that it is not only possible to distinguish between semantic categories, but that a model can even learn to directly encode which word a participant is reading. These encoding analyses provide an interesting evaluation paradigm for the cognitive plausibility of computational representations of language \cite{abnar2017,anderson2017,bulat2017,xu2016, fyshe2014}.

 Recent studies including natural speech find more complex and widespread activation patterns which indicate that fine-grained categorical and topological differences observed for controlled language stimuli cannot be generalized \cite{huth2016natural,hamilton2018revolution}. The results of computational studies using brain data from humans processing full sentences \cite{pereira2018} and even full stories \cite{Jain2018,Dehghani2017,Brennan2016,Wehbe2014} are hard to generalize because the interpretation of language--brain modelling experiments strongly depends on the chosen evaluation metric \cite{beinborn2019} and the differences between models \cite{Gauthier2018}. We therefore also choose to simplify the encoding task and apply representational similarity analysis which makes it possible to compare relations between concepts in computational and cognitive representations more directly \cite{kriegeskorte2008}. 

% \begin{figure}
% \centering
%     \includegraphics[scale=0.4]{figures/StimuliPresentationPereira.png}
%     \caption{Three context paradigms for the concept \textit{bird}. \textsc{Sentence} is shown on top, \textsc{picture} in the middle, and \textsc{word cloud} in the bottom. The example is extracted from Figure 3 in \newcite{pereira2018}.}
%     \label{fig:stimuli}
% \end{figure}

\begin{figure}
\centering
    \includegraphics[scale=0.35]{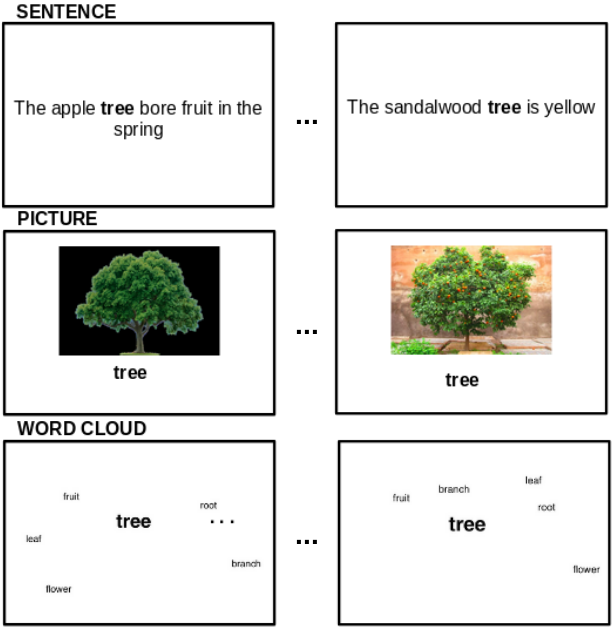}
    \caption{Three context paradigms for the concept \textit{tree}. \textsc{Sentence} is shown on top, \textsc{picture} in the middle, and \textsc{word cloud} in the bottom. The examples are extracted from the stimuli set that was used by \newcite{pereira2018} during the collection of fMRI data.}
    \label{fig:stimuli}
\end{figure}

\subsection{Concreteness of Words}
Concreteness describes the extent to which a word can be embodied by perceptual experiences \cite{walker1999}. Concrete words refer to concepts that are easily perceivable by the senses, %look up the definition again and add a source
for example, a \textit{banana} has a recognizable outlook, feel and taste. Abstract concepts describe theoretical concepts that cannot be directly grounded in perception, for example \textit{democracy}. Psycholinguistic research on lexical access indicates that concrete words are usually processed more rapidly and accurately than abstract concepts \cite{kroll1986}. However, some patients show a reversed pattern \cite{breedin1994}. This hints at different underlying brain representations for abstract and concrete concepts and led to the development of two main theories.

%Psychological theories concerning these brain representation include the dual code theory \cite{paivio1991} and the context availability theory \cite{schwanenflugel1983}. Both theories assume concepts are stored verbally (concerning linguistic relations between related concepts). However, the dual coding theory implies concrete concepts are simultaneously stored non-verbally (according to their perceptual information), while the context availability theory attributes the effects to concrete concepts having richer semantic relations. This is supported by the fact that the recognition of abstract concepts is improved in a supporting context \cite{schwanenflugel1988}. 
The dual coding theory implies that all concepts are stored verbally (as linguistic relations between related concepts) in the brain, while only concrete concepts are simultaneously stored non-verbally (according to their perceptual information; \citealt{paivio1991}). In contrast, the context availability theory disregards non-verbal storage and attributes the effects to concrete concepts having richer semantic relations \cite{schwanenflugel1983}. It has been shown that the recognition of abstract concepts is improved in a supporting context \cite{schwanenflugel1988}. 

Computational representations of words can be based on textual or visual distributional characteristics. Several researchers assume that visual representations are beneficial for modeling concrete concepts and textual representations are better at reflecting relations between abstract concepts \cite{Bruni2014,hill2014,lazaridou2015, beinborn2018}. Recent encoding analyses led to inconsistent results. \newcite{bulat2017} find that visual representations are better for predicting brain activation patterns of concrete concepts, whereas \newcite{anderson2017} do not find a difference between modalities.
% Recent experimental analyses of this hypothesis lead to inconsistent results. \newcite{bulat2017} find that visual representations are better for modelling concrete concepts, whereas \newcite{anderson2017} does not find a difference between modalities.  % ==> needs to be clearer that they're encoding and not testing how good computational representations are in a different manner
Interestingly, the stimuli used in the two studies differ (words + drawings vs words, respectively). In the current work, we analyze the same concepts presented in different presentation contexts to further explore the influence of experimental paradigms.
%\todo{Improve figure for cameraready!}

\section{Data}
In this section, we provide details on the fMRI dataset and the preprocessing methods.
\subsection{FMRI Dataset} 
\newcite{pereira2018} presented 180 concepts (128 nouns, 22 verbs, 23 adjectives, 6 adverbs, 1 function word) to 16 participants and measured their fMRI response. Participants were instructed to think of the target concept in terms of the corresponding context. In line with \newcite{pereira2018}, we use the term concept instead of word to account for this particular experimental setting.  
\paragraph{Context} The concepts were presented to the participants within varying context paradigms: \textsc{sentences} with the target concept in a bold font; \textsc{pictures} presented alongside the target concept; and \textsc{word clouds} with the target concept in the center, surrounded by semantically related words. Figure \ref{fig:stimuli} provides examples for the different presentation contexts for the concept \textit{tree}.

\paragraph{Concreteness} The concepts are annotated with concreteness ratings \cite{brysbaert2014}. We categorize concepts as concrete
if their concreteness score lies at least half a standard deviation above the mean, and abstract if it lies at least half a standard deviation below the mean.\footnote{mean = 3.49, std = 1.08} This results in 69 concrete and 63 abstract concepts. 
\begin{figure*}
    \subfloat{\includegraphics[height=1.47in]{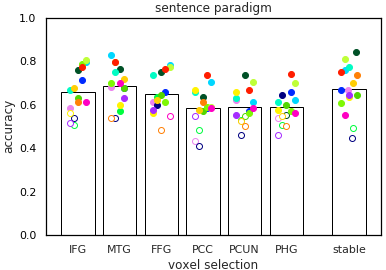}}
    \hspace{0.5em}
    \subfloat{\includegraphics[height=1.47in]{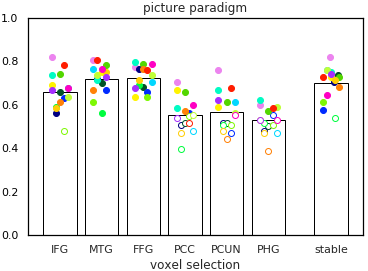}}
    \hspace{0.5em}
    \subfloat{\includegraphics[height=1.47in]{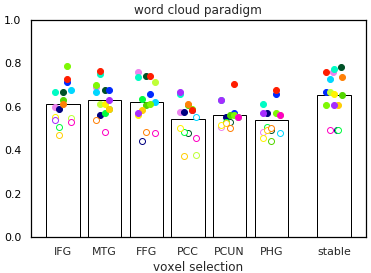}}
    \newline
    \subfloat{\includegraphics[scale=0.38]{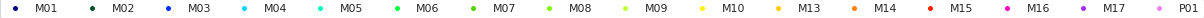}}
    \caption{Accuracy scores for the decoding analysis for different regions. Each color represents a subject. Significant results are indicated by filled circles. The abbreviations stand for the regions Inferior Frontal Gyrus (IFG), Middle Temporal Gyrus (MTG), FusiForm Gyrus (FFG), posterior cingulate (PCC), precuneus (PCUN), and parahipocampal gyrus (PHG), and for the stable voxel selection across paradigms (stable).}
    \label{fig:classification_results_roi_stable}
\end{figure*}
\subsection{Pre-processing}
The fMRI dataset has already undergone standard pre-processing. During the experiment by \newcite{pereira2018}, each concept has been presented to the participants four to six times in each context paradigm. The scans have been averaged over these instances to obtain a generalized concept scan for each paradigm.  

FMRI scans are commonly represented by a matrix of voxels. A voxel can conceptually be understood as a 3-dimensional fixed-size pixel in a brain scan.\footnote{In the data by \newcite{pereira2018}, the dimensions of these voxels are 2x2x2 mm and the number of voxels per participant ranges from 145,303 to 212,742.}
As not all voxels are expected to represent relevant information, computational analyses are usually performed on a subset of the total voxels \cite{Wehbe2014,Brennan2016}. In our analysis, we compare different subsets by selecting regions of interest, determining stable voxels, and performing a searchlight analysis.

\paragraph{Regions of Interest (ROI)} A common reduction method restricts the brain response to voxels that fall within a pre-selected set of brain regions. For the discrimination between abstract and concrete concepts, the regions of interests found in previous studies vary  \cite{mestres2009,tettamanti2008,wallentin2005,wise2000}. A large meta-analysis finds abstract concepts elicit more activation in the inferior frontal gyrus (IFG) and middle temporal gyrus (MTG), and concrete concepts elicit more activation in the posterior cingulate (PCC), precuneus (PCUN), fusiform gyrus (FFG), and parahippocampal gyrus (PHG) \cite{wang2010}. All discriminating regions were located in the left hemisphere. For our analyses, we select these regions.
%Voxels are aligned to AAL regions based on a mapping that was provided with the data. 

 \paragraph{Stable Voxels} Selecting stable voxels is a more data-driven approach to voxel selection. This strategy aims to select voxels that consistently capture the representation of a concept. The concepts have been presented to the participants in three paradigms, which require different processing steps, e.g., \textsc{sentences} require reading skills to combine characters into a meaning representation, \textsc{pictures} require visual processing to combine pixels into an image representation, and \textsc{word clouds} require spatial understanding to parse the word cloud. To approximate the joint underlying conceptual representations, we determine 500 voxels with the most stable activation pattern across presentation paradigms. This selection method is inspired by the pre-processing applied in \newcite{Mitchell2008} to detect stable voxels across experimental runs.

\paragraph{Searchlight} \newcite{KriegeskorteSearchLight} proposed another type of data-driven voxel selection. They only analyze a small sphere of neighbouring voxels and assign the result to the center voxel of the sphere. The sphere moves through the entire brain, so that each voxel is used as the center once. We use a 4mm radius resulting in a sphere of 33 voxels (except for the margins of the brain) for each analysis step. % \footnote{The searchlight analyses is conducted on the native space of each participant} 

\section{Discriminative Analyses}
Discriminative analyses such as decoding and clustering operate only on the fMRI scans. The algorithms identify hubs in the high-dimensional data and the output is evaluated with respect to our linguistic category concreteness.
Decoding %(or classification) 
is a common analysis method for fMRI data \cite{fMRIhandbook}. The data is split into training and test data and the algorithm learns to find a hyperplane separating the representations in the training data according to the corresponding class annotations (concrete/abstract in our case). Based on this separation, the class for the representations in the test data is predicted.
In clustering, the goal is to automatically find a separation of the data into $k$ homogeneous groups without any prior class bias. Our experimental code is based on the fMRI evaluation framework by \newcite{beinborn2019}. For classification and clustering, we use standard algorithms from the python library \textit{scikit-learn} \cite{scikit-learn}. 

%TODO: for camera-ready: add github link to our code
\begin{figure}
\centering
    \subfloat[\textsc{sentence}]{\includegraphics[height=1in]{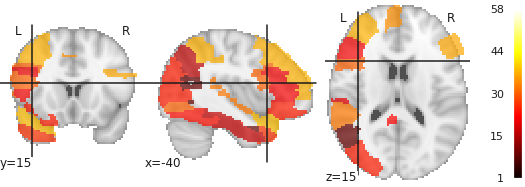}}
    \newline
    \subfloat[\textsc{picture}]{\includegraphics[height=1in]{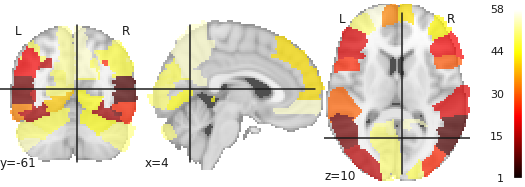}}
    \newline
    \subfloat[\textsc{word cloud} ]{\includegraphics[height=1in]{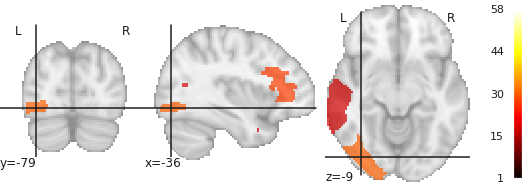}}
    \caption{Decoding results of the searchlight analyses for the three context paradigms. We highlight areas with an average decoding accuracy $\geq 0.52$. The colors indicate the average rank of the area (relative to all brain areas) over all subjects. 
    %A darker color indicates a higher average ranking, i.e., a higher encoding accuracy compared to other areas. 
    }
    \label{fig:classification_results_searchlight}
\end{figure}
\subsection{Decoding}
 We split the dataset into 11 folds of 12 concepts and perform cross-validation using a support vector machine with hyperparameter settings as recommended by \newcite{song2011}.\footnote{Radial basis function kernel with the gamma coefficient set to scale.} 

\paragraph{ROI \& Stable Voxel Results}
The accuracy scores for the decoding task for different regions are visualized in Figure \ref{fig:classification_results_roi_stable}. The three subplots refer to the three presentation paradigms \textsc{sentences} (left), \textsc{pictures} (middle), and \textsc{word clouds} (right). 
Significance of results is determined by comparing to the average score of a baseline on randomly permuted labels which was repeated 1,000 times.\footnote{The significance level was set to $\alpha$ $<$ 0.05. Note that the smallest p-value obtainable with this distribution is 0.001.} Filled circles visualize subjects with significant decoding accuracy, empty circles indicate insignificant results.  

We can see that the model learns to discriminate between the two categories for more than 75\% of the subjects in the regions IFG, MTG, and FFG. This finding is consistent across context paradigms with the best results obtained for the \textsc{picture} paradigm. As in previous work, the variance between subjects is quite high with accuracy values over 80\%. The results for the remaining regions are close to random. It should be noted, that some of these regions include less voxels which might degrade the expressivity of the model. The data-driven approach to select stable voxels across paradigms yields accuracy scores that are competitive with the best regions.  

\paragraph{Searchlight results} In order to abstract from the size of a region, we conduct the searchlight analysis.
%sphere with a constant size of 33 voxels (except for the margins of the brain)
For each sphere, we calculate the decoding accuracy. To compare the sphere results across participants, we calculate the average decoding accuracy of all spheres within a brain area.\footnote{Brain areas are determined by a mapping according to the automated anatomical labeling atlas as indicated by \newcite{pereira2018}.} We then rank the areas from highest to lowest accuracy for all participants. In Figure \ref{fig:classification_results_searchlight}, we visualize the average ranks of the areas with a decoding accuracy $\geq 0.52$. 

We identify a larger number of decoding regions in the \textsc{picture} context than in the \textsc{sentence} and \textsc{wordcloud} context. Strikingly, we consistently obtain the highest ranks for the middle temporal gyrus (MTG) in all context paradigms.
%9.88 in the \textsc{sentence} paradigm, 7.63 in the \textsc{picture} paradigm, and 18.94 in the \textsc{word cloud} paradigm. 
However, in the paradigms that only use linguistic stimuli (i.e., \textsc{sentence} and \textsc{word cloud}), the left MTG is ranked on top, while it is the right MTG for the \textsc{picture} context. In line with previous work stating that linguistic processing mainly elicits activity in a left lateralized network (e.g., \citealt{frost1999}; \citealt{knecht2000}), the majority of the highly ranked areas in the linguistic pardigms are located in the left hemisphere. 

% \todo{If time: Provide a confusion matrix for the best subject to see how many concrete and abstract concepts are correctly/incorrectly classified so that we get a better sense of the quality.}
% \todo{Are there any concepts that are misclassified across several of the "good" subjects?}
\subsection{Clustering}
We have seen that it is generally possible to distinguish between concrete and abstract concepts to a certain extent. This indicates that the representational patterns for the two categories differ in at least one dimension. The clustering analysis provides further information of the concreteness distinction could be considered as a "natural" class for the fMRI representations of semantic concepts.
 We set $k$ to 2 and run a $k$-means algorithm to categorize the fMRI representations for all concepts. We then analyze to which extent these two clusters correspond to our abstract/concrete distinction. 
% of concepts and compare the clusters to the concreteness categories using the Adjusted Rand Index \cite{hubert1985}. \todo{If space add formula, if not add 1-2 descriptive sentences.}
\paragraph{Results}
Table \ref{tbl:clustering} provides the results of the clustering averaged over subjects. We investigate the proportion of abstract concepts in each cluster. For readability, we only report the first cluster and only the three regions that worked well in decoding.\footnote{The tendencies remain consistent for the other cluster and regions. %\todo{see supplementary material?}
} 
We see that across regions and paradigms the proportion of abstract concepts in the cluster always reflects the proportions of the dataset (0.48). This shows that concreteness is not the most prevalent category for grouping the fMRI patterns of the stimuli. Another latent factor seems to be more dominant. A first manual inspection of the clustering results did not yet raise a hypothesis regarding the characteristics of this latent factor.

We have seen that the concreteness distinction can to a certain extent be decoded from the fMRI signal. However, this does not mean that it is the predominant distinctive feature for the stimuli. In order to get a better idea about the structure of the representations, we conduct relational analyses. 

\section{Relational Analyses}
While discriminative analyses examine whether two classes of concepts can generally be separated, relational analyses provide more information on the representations of individual concepts. 
%this is such a gorgeous sentence!:
They indicate to which extent computational models simulate the relations between concepts that are observed in the human cognitive signal. 
\paragraph{Computational Representations}
We use \textsc{Glove} embeddings as a textual representation of the concepts \cite{glove} because they are recommended for psycholinguistic experiments \cite{pereira2016}.\footnote{We used the 300-dimensional model trained on 42 billion tokens available here: \url{https://nlp.stanford.edu/projects/glove/}}
We generate visual representations of each concept by running a pre-trained ResNet from the \textit{keras} library on the six images representing the concept in the \textsc{picture} paradigm and averaging over the six resulting image vectors. 

\subsection{Encoding}
For encoding, a linear regression model learns to predict the corresponding fMRI pattern for a given computational representation of a concept in the training data. We split the data into 11 folds of 12 concepts and perform cross-validation. The trained model is evaluated by predicting the fMRI pattern for the concepts in the test data. We calculate the pairwise accuracy based on the implementation in \newcite{beinborn2019} using the \textit{single} match definition and cosine distance. This evaluation metric tests if the model prediction for a test concept, e.g. \textit{tree}, is more similar to the observed scan for \textit{tree} than to the observed scans for another concept. This comparison is performed for all concepts in the dataset and the accuracy for the test concept is averaged. Accuracy is calculated separately for abstract and concrete concepts.
\begin{table}
\small
\begin{tabular}{llll} 
\toprule
Region & \textsc{sentence} &  \textsc{picture} &  \textsc{word cloud}  \\
\midrule
IFG    &    0.49 ($\pm$ 0.05)&    0.46 ($\pm$ 0.11)&  0.49 ($\pm$ 0.05) \\
MTG     &   0.48 ($\pm$ 0.07) &   0.45 ($\pm$ 0.14) &   0.46 ($\pm$ 0.05)  \\
%PCC &  0.47 ($\pm$ 0.07) &   0.48 ($\pm$ 0.05) &   0.45 ($\pm$ 0.06\\
%PCUN    &   0.49 ($\pm$ 0.06) &   0.48 ($\pm$ 0.07) & 0.49 ($\pm$ 0.05\\
FFG     &   0.50 ($\pm$ 0.09)	&	0.49 ($\pm$ 0.07)	&   0.48 ($\pm$ 0.05)\\
%PHG &   0.50 ($\pm$ 0.07)	&	0.47 ($\pm$ 0.05)	&	0.45 ($\pm$ 0.06)\\
Stable &    0.52 ($\pm$ 0.13)   &   0.45 ($\pm$ 0.15) & 0.47 ($\pm$ 0.11)\\
\bottomrule
\end{tabular}
\caption{Proportion of abstract concepts in the first cluster learned for  different regions averaged over all subjects. The results for the second cluster are similar. Note that the results reflect the proportion of abstract concepts in the whole dataset (0.48).}
\label{tbl:clustering}
\end{table}

\begin{figure*}
\centering
    \subfloat[\textsc{sentence}]{\includegraphics[scale=0.24]{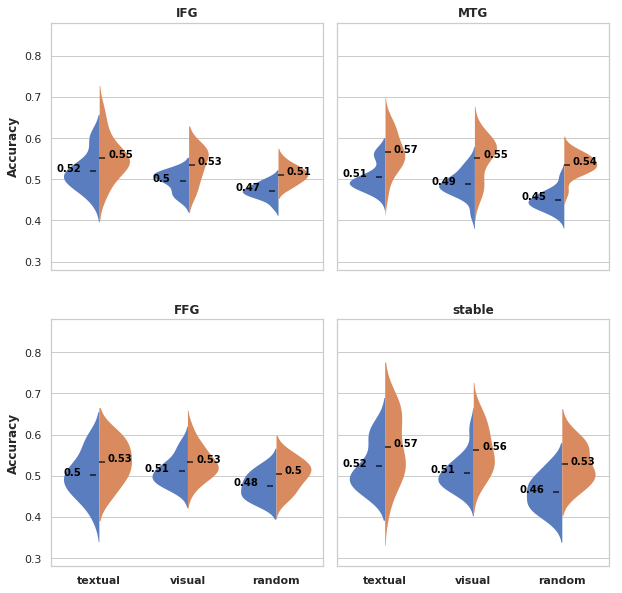}}
    \hfill
    \subfloat[\textsc{picture}]{\includegraphics[scale=0.24]{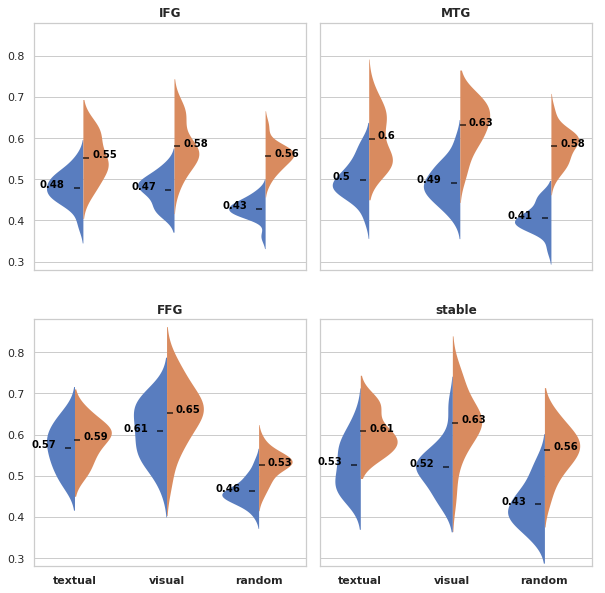}}
    \hfill
    \subfloat[\textsc{word cloud}]{\includegraphics[scale=0.24]{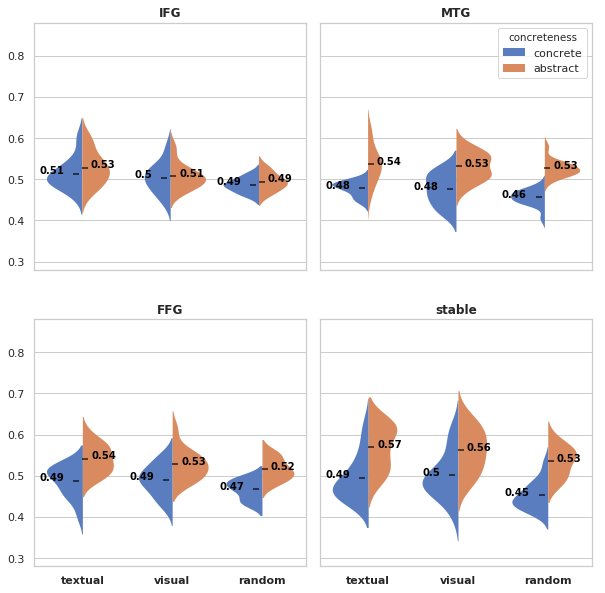}}

    \caption{Accuracy results for the encoding analysis as a density estimation over all subjects for the Inferior Frontal Gyrus (IFG), Middle Temporal Gyrus (MTG), FusiForm Gyrus (FFG), and the selected stable voxels (Stable). Accuracy is calculated separately for abstract and concrete concepts. The stimuli are encoded by the model using textual, visual, and random computational representations.}
    \label{fig:encoding}
\end{figure*}
\paragraph{Results}
Figure \ref{fig:encoding} shows the encoding accuracy for all participants with significant results in decoding as violin plots. For clarity, we only plot the regions for which we obtained good decoding results in the previous analysis. The average accuracy is calculated separately for concrete and abstract concepts. We compare the encoding results for the textual and visual representations to randomly initialized vectors. The random results are averaged over 1,000 different initializations.

We see that all encoding results are close to random and we do not find a strong difference between textual and visual representations. The accuracy is slightly higher in the \textsc{picture} paradigm for the FFG area, but the variance of the results is also higher. Interestingly, we observe the pattern that abstract concepts seem to be slightly easier to encode than concrete concepts. The effect is very small, but it is consistent across all experimental settings. However, we even observe it for the random representations which indicates that the effect cannot be explained linguistically, because the random representations do not distinguish between concrete and abstract concepts. We have not yet converged on a convincing hypothesis for explaining this effect. Given the small dataset, it might just be a distributional artifact. 

\subsection{Representational Similarity Analysis}
For conducting representational similarity analysis \cite{kriegeskorte2008}, we do not need to learn an intermediate mapping model. Instead, the relations between the fMRI activation vectors are directly compared to the relations that can be observed in the computational representations of the stimulus \cite{anderson2016}. % As commonly performed, 
We measure the relations between vectors within cognitive and computational representations by %Pearson correlation 
cosine distance and compare between fMRI data and the computational representations using Spearman correlation.
\begin{figure*}[h]
\centering
    \subfloat[\textsc{sentence}]{\includegraphics[scale=0.24]{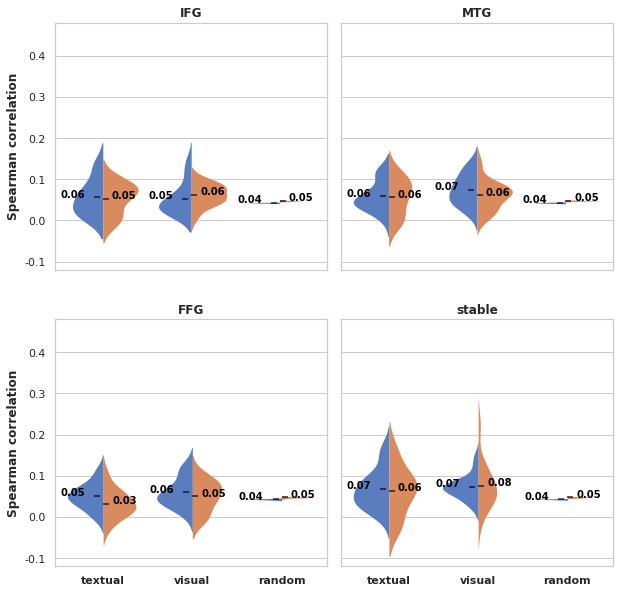}}
    \subfloat[\textsc{picture}]{\includegraphics[scale=0.24]{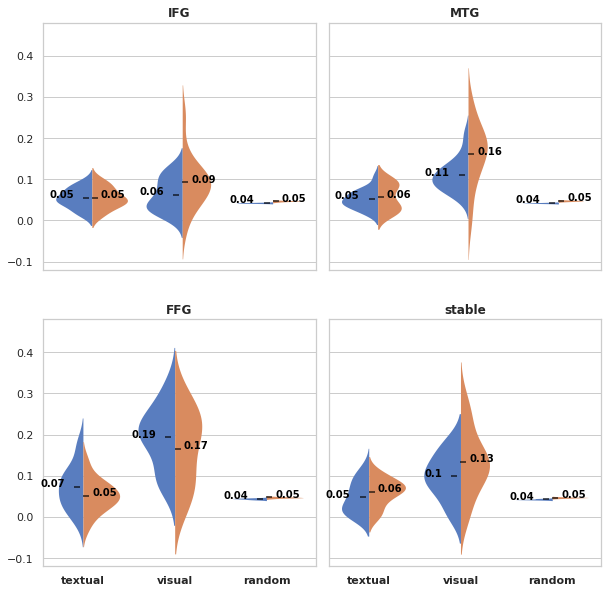}}
    \subfloat[\textsc{word cloud}]{\includegraphics[scale=0.24]{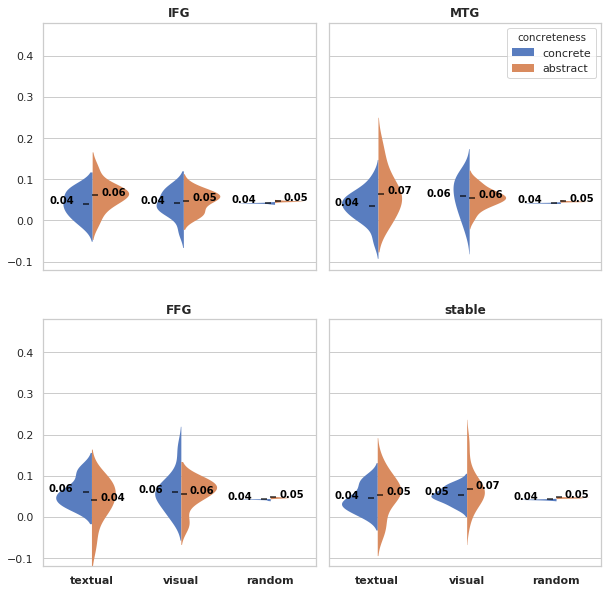}}
    \caption{Spearman correlation between computational representations of the stimuli (textual, visual, random) and the fMRI activation patterns in the Inferior Frontal Gyrus (IFG), Middle Temporal Gyrus (MTG), FusiForm Gyrus (FFG), and the selected stable voxels (stable). Correlation is calculated separately for abstract and concrete concepts.
    \todo{Adjust caption to use the term Representational Similarity Analysis!}}
    \label{fig:rsa}
\end{figure*}

\paragraph{Results} Figure \ref{fig:rsa} visualizes the results for the Spearman correlation in violin plots for participants with significant results in decoding. We can see that the correlation values are very low in all cases. 
%As in encoding, the results are slightly higher for visual representations in the \textsc{picture} paradigm, but with a high variance. 
The higher variance and slightly higher accuracies of the visual representations could be related to the perception of object form in the image stimulus, rather than semantic processing of the context (which has been shown to occur in the FFG by e.g., \citealt{whatmough2002}).

The results for the encoding task and the representational similarity analysis indicate that the way relations between words are modeled in the computational representations differs from the relations observed in human language processing. In contrast to previous analyses, we do not find a difference between the modelling quality of concrete and abstract concepts in textual and visual representations. While the decoding analysis supported the assumption that words can be categorized based on the concreteness distinction, the relational analyses reveal that the problem is more complex. 
\section{Discussion}
% \todo{This currently reads a bit too negative, if possible find a slightly more positive framing }

We have seen that different analysis techniques have a strong influence on the interpretation of the results. Based on the decoding results, it can be tempting to draw too simplistic conclusions regarding the categorizability of words, which are not supported by the relational analyses.
Our comparison only represents a first exploratory analysis. In order to come to more stable conclusions, we plan to explore a wider range of linguistic categories and conduct more fine-grained error analyses. We believe that it is important to focus less on the significance of results in a single experimental paradigm and instead explore a range of analysis techniques to test alternative explanations. 
 
The robustness of results can be attacked by conducting sensitive sanity checks and comparing to more reasonable baselines. Seemingly positive results can sometimes even be obtained with a scrambled signal or turn out to be not significantly different from results obtained by a carefully fine-tuned random baseline involving no linguistic knowledge. The interpretation of the results is further complicated by the problem that fMRI scans produce a high-dimensional and noisy signal that needs to be pre-processed using several statistical correction techniques. Such pre-processing steps have a strong influence on the results of an fMRI study \cite{strother2006} and their effect on modelling analyses has not yet been sufficiently studied. Furthermore, the high inter-subjective variance in fMRI analyses poses an additional challenge. 

In line with the idea by \newcite{hamilton2018revolution}, we have seen that the kind of context in which a word is presented also plays an important role. 
% The searchlight analysis seems to line with previous findings by \newcite{huth2016natural}, we see that more complex paradigms elicit a more elaborate, widespread activation in the brain. However no searchlight analyses for the other analysis were performed and their accuracies were very low
%because concepts can be used in both abstract and concrete contexts. 
To develop a better understanding of the differences and commonalities of visual and textual semantic processing, grounded multimodal language models are a promising development \cite{baroni2016}.
We hypothesize that even within one modality, context can determine whether a word is experienced as more abstract or concrete. Contextualized language models like ELMO \cite{peters2018} and BERT \cite{devlin2018} might therefore provide more suitable representations.  
% talk about inter-subjective differences?

From our detailed analysis, we obtain the impression that human concept representations are more fluid than dichotomous categories can capture. We are dealing with high-dimensional data and one-dimensional explanations can only provide one perspective. Recent analyses on the interpretability of computational language models attempt to decode linguistic features such as syntactic structure from computational models \cite{alishahi2019}. The challenges for interpretation are similar: the observation that we can decode a category from the signal does not necessarily imply that the signal is structured according to the category. We believe that embracing the many shades of grey between concepts will lead to more realistic models of cognitive processing because natural language is fluid, ambiguous, and multi-faceted. 
 
\section{Conclusion}
We analyzed to which extent the distinction between concrete and abstract concepts can be extracted from fMRI data using a range of discriminative and relational analysis methods. We find that the distinction can be decoded from the signal with an accuracy significantly above chance, but it is not found to be a relevant structuring factor in the clustering and relational analyses. 

%i love this part
We do not discourage the use of dichotomous categories for analysis because they might be a useful explanatory simplification for linguistic phenomena. However, our exploratory analyses indicate that researchers should be aware that the cognitive processing structure is more complex. Language meaning is constantly evolving, and susceptible to manipulation. In the long run, accepting the fluidity of concept representations seems more fruitful than artificially mapping the high-dimensional probabilistic representations back into binary categories.

\section*{Acknowledgement}
The work by L. Beinborn was funded by the Netherlands Organisation for Scientific Research (NWO), through a Gravitation Grant (024.001.006) to the Language in Interaction Consortium.

\bibliography{brainlang}
\bibliographystyle{acl_natbib}
\end{document}